\documentclass{esannV2}
\usepackage[dvips]{graphicx}
\usepackage[latin1,utf8]{inputenc}
\usepackage{amssymb,amsmath,array}

\usepackage{multirow}
\usepackage{adjustbox}
\usepackage{url}
\usepackage{latexsym}
\usepackage{array}
\usepackage{todonotes}
\usepackage{subfigure} 
\usepackage{algorithm}
\usepackage{algorithmic}
\usepackage{makecell}

\begin{document}
\title{Attention-based Ingredient Phrase Parser}

\author{Zhengxiang Shi, Pin Ni, Meihui Wang, To Eun Kim and Aldo Lipani
%
%
\vspace{.3cm}\\
%
University College London \\
Gower St, London - United Kingdom
%
}

\maketitle

\begin{abstract}
As virtual personal assistants have now penetrated the consumer market, with products such as Siri and Alexa, the research community has produced several works on \textit{task-oriented dialogue tasks} such as hotel booking, restaurant booking, and movie recommendation.
Assisting users to cook is one of these tasks that are expected to be solved by intelligent assistants, where ingredients and their corresponding attributes, such as name, unit, and quantity, should be provided to users precisely and promptly. However, existing ingredient information scraped from the cooking website is in the unstructured form with huge variation in the lexical structure, for example, “1 garlic clove, crushed”, and “1 (8 ounce) package cream cheese, softened”, making it difficult to extract information exactly.
To provide an engaged and successful conversational service to users for cooking tasks, we propose a new ingredient parsing model that can parse an ingredient phrase of recipes into the structure form with its corresponding attributes with over 0.93 F1-score. 
Experimental results show that our model achieves state-of-the-art performance on AllRecipes and Food.com datasets.

\end{abstract}

\section{Introduction}
\begin{figure}[!t]
  \centering
  \includegraphics[width=0.81\textwidth]{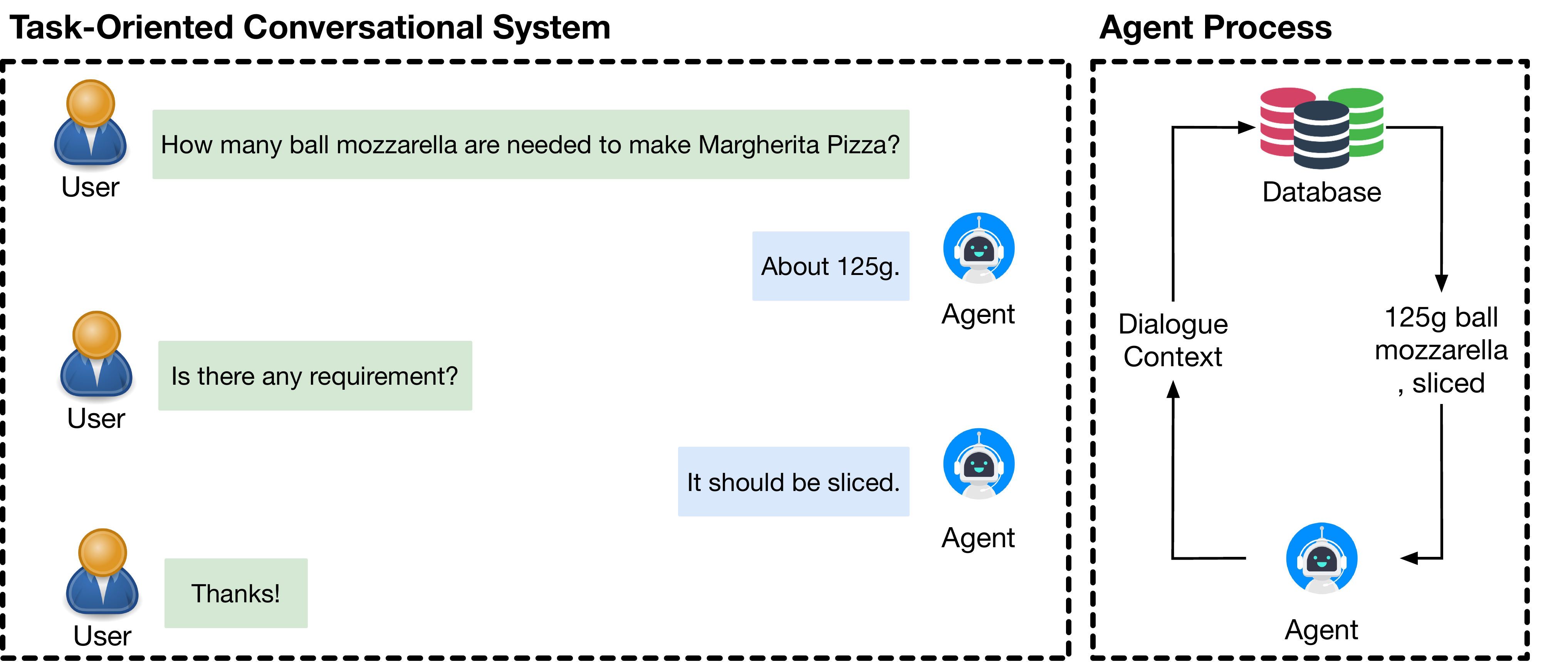}
  \caption{Applications of our ingredient parser model for conversational systems: The user may ask for details of ingredients (left side). The agent will then retrieve relevant recipes or cooking information from the pre-built database based on the dialogue states and contexts (right side).}
  \label{fig:example}
\end{figure}

There are few things so fundamental to our life as food, whose consumption is intricately linked to our health, our feelings and our culture.
With the rapid development of science and technology, conversational AI has been a long-standing area of exploration in the research community~\cite{shi2022stepgametexts,Shi2022learning,kim2022multi} and has now penetrated in both academia and industries with products such as Microsoft Cortana and Amazon Alexa. 
Recently, 
researchers work on
integrating cooking tasks into conversation systems
with the target to assist customers to complete everyday tasks~\cite{ramoscondita}.
To assist users with cooking tasks, fine-grained information about each recipe, such as cooking processes, utensils, nutritional profile, dietary style, and ingredient details, are needed. In particular, the ingredients details of a recipe typically contain attributes such as quantity, temperature, and processing state.
Moreover, ingredient information itself can have use cases such as food pairing, flavour prediction, nutritional estimation, cost estimation and cuisine prediction.
In Figure~\ref{fig:example}, we show an example of a conversation where the intelligent agent is assisting a user to cook. Here we can see that the quantity and other information about the recipe are necessary to respond to users' requests. 

Although there are several datasets for the cooking domain, such as Recipe\-1M+~\cite{marin2019recipe1m+} and RecipeNLG~\cite{bien2020recipenlg}, and a plethora of websites, such as AllRecipes and WikiHow, providing plenty of human-readable recipes, there is a lack of structured data about recipes useful to enable complex queries. 
Specifically, existing ingredient information and their corresponding attributes are in an unstructured form with huge variations, for example, “1 garlic clove, crushed”, and “1 (8 ounce) package cream cheese, softened”, making it difficult to extract precise attribute information.
As shown on the right side of Figure~\ref{fig:example}, the available ingredient information that we can extract from existing databases is all in the unstructured form.
Addressing this problem needs the implementation of natural language processing algorithms that identify relevant attributes (quantity, unit, temperature, processing state, etc.) from ingredient phrases.
Different from common machine learning tasks, this ingredient phrase parsing task requires high performance to provide users with more engaging and satisfactory conversations.
Diwan et al.~\cite{diwan2020named} presented two dataset, AllRecipes and Food.com, with $8\,800$ annotated ingredient phrases based on the RecipeDB~\cite{batra2020recipedb}.
However, there is no ingredient parsing model with publicly available code on these two datasets.
To this end, we propose a novel attention-based neural network model for ingredient parsing.
Experimental results demonstrate that our model can achieve state-of-the-art performance on AllRecipes and Food.com datasets. 

\section{Related Work}

The neat definition of cooking tasks and the need to develop intelligent agents 
has recently attracted much interest from the research community.
%
%
Marin et al.~\cite{marin2019recipe1m+} proposed Recipe1M+ dataset, a large-scale, structured corpus of over one million cooking recipes (including cooking instructions and ingredients) and $13$
million food images. Based on this dataset, they trained a neural network to learn a joint embedding of recipes and images for the image-to-recipe retrieval task.
Bien et al.~\cite{bien2020recipenlg} introduced RecipeNLG, a dataset of cooking recipes, built upon Recipe1M+. 
RecipeDB~\cite{batra2020recipedb} presents a structured, annotated dataset of over $118{\small,}171$ recipes, which are composed of $23{\small,}548$ ingredients.
The recipes have been classified into cuisines represented by $26$ geo-cultural regions that span $6$ continents and containing $74$ countries with ingredients grouped into $29$ categories. 
Diwan et al.~\cite{diwan2020named} presented a labelled dataset of $8{\small,}800$ ingredient phrases based on the RecipeDB~\cite{batra2020recipedb}, divided into training and testing sets. 
Unlike Marin et al. and Bien et al.~\cite{marin2019recipe1m+,bien2020recipenlg}, the authors manually tagged ingredient phrases into 8 categories. 
Diwan et al.~\cite{diwan2020named} proposed a method to map ingredient phrases into a structured format with their corresponding attributes via the utilization of clustering and Named Entity Recognition (NER) based on the Stanford NER~\cite{finkel2005incorporating}. However, the code for this method is not available.

\section{Method}
\begin{figure}[!t]
  \centering
  \includegraphics[width=0.80\textwidth]{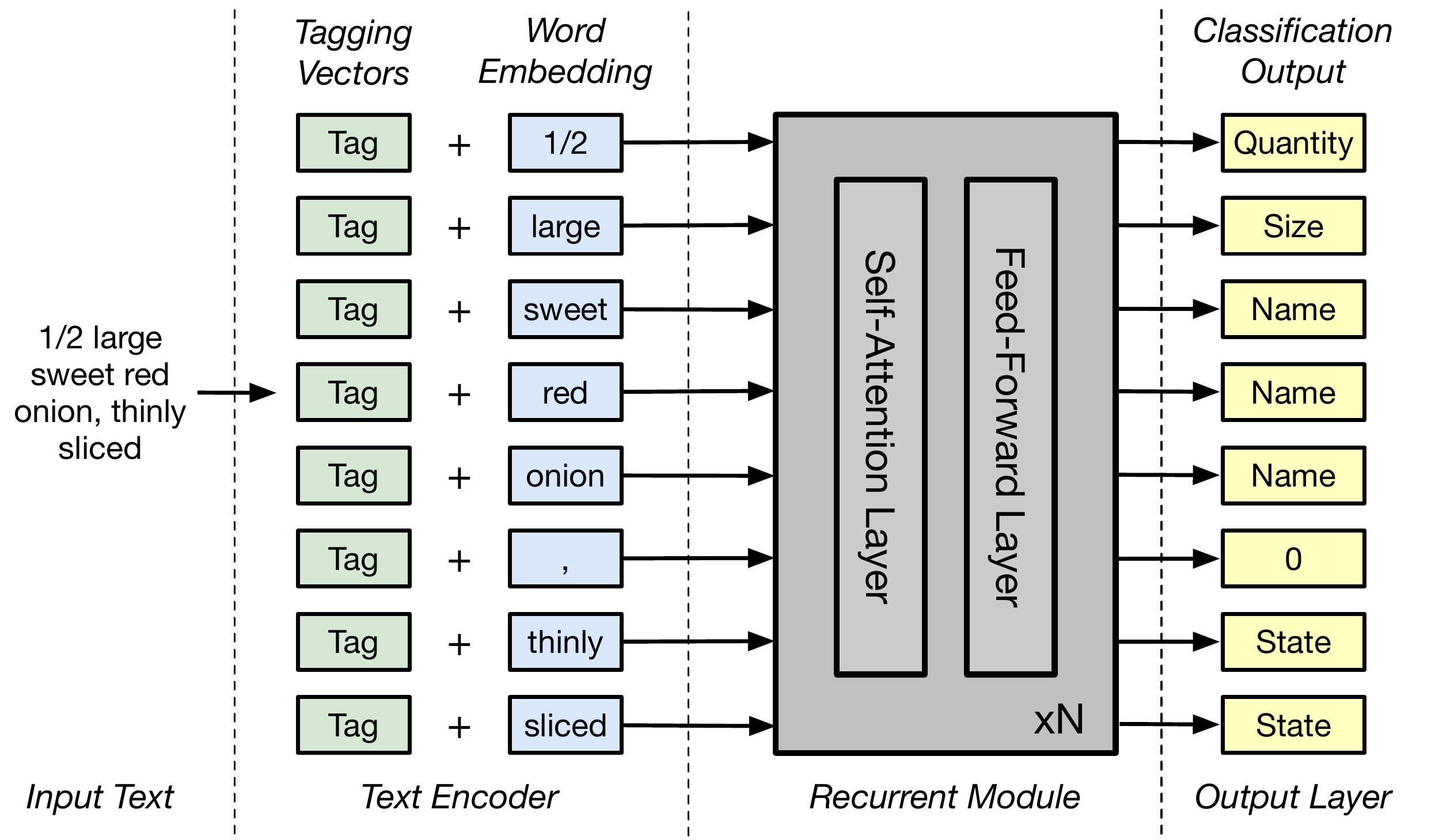}
  \caption{The model architecture. Each green cell represents the encoded vector for its predicted POS tags and each blue cell stands for the encoded vector for this token. The output layer will predict one class for each token (yellow cell). N represents the number of layers in the recurrent module.
  }
  \label{fig:model}
\end{figure}

In this section, we introduce the proposed ingredient parsing model, as shown in Figure~\ref{fig:model}.
The model consists of three major components: a text encoder, an attention-based recurrent module, and an output layer. 
We first revisit the definition and notations about the widely used attention mechanism~\cite{bahdanau2014neural,ni2020natural,ni2021hybrid,ni2021staresgru} and then introduce the details of our proposed ingredient parsing model.

\paragraph{\textbf{Attention Mechanism.}}

Given a query vector $x$ and a sequence of context vectors $\{y_j\}^{K}_{j=1}$, the attention mechanism first computes the matching score $s_j$ between the query vector $x$ and each context vector $y_j$. Then, the attention weights are calculated by normalizing the matching score: %
$a_j = \frac{exp(s_j)}{\sum_{j=1}^{K}exp(s_j)}$.
The output of an attention layer is the attention weighted sum of the context vectors:
$Attention(x, {y_j}) = \sum_{j} a_j \cdot y_j$.
Particularly, the attention mechanism is called self-attention when the query vector itself is in the context vectors $\{y_j\}$. 

\paragraph{\textbf{Text Encoder.}}

We first pre-process the ingredient phrases by tokenizing them using the NLTK toolkit~\cite{bird2004nltk}, and converting the tokens to lower case. 
Given an ingredient phrase, $\{w_i\}_{i=1}^{s}$, where $w_i$ is the $i$-th token and $s$ is the number of tokens in the sequence, we use the text encoder to encode the sequence of tokens into a sequence of vectors, $\{u_i\}_{i=1}^{s}$, where $u_i$ is the $d$-dimension vector representing the $i$-th token. 
In this paper, we encode all tokens as 300-dimensional pre-trained word embeddings from Glove~\cite{pennington2014glove}. For each token out of the vocabulary, we will encode it as a trainable 300-dimensional random vector. 
To provide additional signals for the model, we also obtain each token's part-of-speech (POS) tag using the NLTK toolkit~\cite{bird2004nltk}. Each tag is then converted into a $d$-dimension vector via a trainable lookup table. Finally, token vectors, $\{u_i\}_{i=1}^{s}$, are summed to their corresponding tag vectors, as shown in Figure \ref{fig:model}.

\paragraph{\textbf{Recurrent Module.}}

The recurrent module consists of $N$ layers, where each layer contains one self-attention module and one feed-forward layer. Each feed-forward layer is further composed of a linear transformation layer, a dropout layer~\cite{srivastava2014dropout} and a normalization layer~\cite{ba2016layer}.  
In each layer, we update the token vectors $\{u_i\}_{i=1}^{s}$ through an attention layer and feed-forward layer.
The output of the current layer will be used as the input of the next layer.
Only the output of the final layer will be fed into the output layer.

\paragraph{\textbf{Output Layer.}}

The output layer is used to map each word embedding $\{u_i\}_{i=1}^{s}$ into a scalar score.
The output layer contains a linear projection layer of trainable parameters and a softmax function.
Then we compute the output logits for each token $u_i$: 
$\hat{s_i} = \text{softmax}(u_i \cdot W)$,
where $W \in \mathbb{R}^{d \times o}$ and $o$ is the number of classes to be predicted. 

\section{Experimental Setup}
\label{sec:experiment}
In this section, we evaluate our model for the ingredient parsing task.
We also present the datasets we used and discuss the baselines. 
Code and data are available at:~\url{https://github.com/ZhengxiangShi/IngredientParsing}.

\paragraph{\textbf{Datasets.}} 

In the following experiments, we will use two datasets, AllRecipes and Food.com datasets, annotated by~\cite{diwan2020named}. The statistics of these two datasets are shown in Table~\ref{table:dataset_statistics}.
\begin{table}[!ht]
\footnotesize
\centering
\begin{adjustbox}{max width=\textwidth}
\begin{tabular}{lccc}
\hline
\multicolumn{1}{c}{Dataset}          & AllRecipes          & Food.com          & Both \\ \hline
Train Set Size                       & 1470                & 5142              & 6612 \\
Test Set Size                        & 483                 & 1705              & 2188 \\ \hline
\end{tabular}
\end{adjustbox}
\caption{Statistics of AllRecipes, Food.com datasets, and their combination.}
\label{table:dataset_statistics}
\end{table}
In these datasets, each ingredient phrase includes ingredients used in the recipe and their corresponding attributes. There are a total of eight types of attributes: 
(1): \textbf{Name}: Name of the ingredient, e.g., salt or pepper; 
(2): \textbf{State}: Processing state of the ingredient, e.g., ground or thawed; 
(3): \textbf{Unit}: Measuring unit of the ingredient, e.g., gram, or, tablespoon; 
(4): \textbf{Quantity}: Quantity of the ingredient, e.g., 1/2;
(5): \textbf{Size}: Portion sizes mentioned, e.g., small or large; 
(6): \textbf{Temperature}: Temperature of the ingredient prior to cooking, e.g., hot or cold; 
(7): \textbf{Dry/Fresh}: Whether the ingredient is dry or fresh, e.g., freshly;
(8): \textbf{Others}: Punctuation in the ingredient phrase, e.g., ",".

\paragraph{\textbf{Baselines.}} 

Diwan et al.~\cite{diwan2020named} claim that their name entity recognition model, which is the only baseline model in the literature, achieves a 0.95 F1-score across all above-mentioned datasets. However, their code and software for this model are not publicly available. After reproducing their results we only manage to obtain a 0.61 F1-score (as shown in Table \ref{table:results1}). Unfortunately, we are not able to justify this discrepancy. Our implementation is based on the interpretation of the description provided in their paper. The authors did not reply when queried about it. 
To also allow the reproducibility of their work, our implementation of this baseline is available in our repository.

\paragraph{\textbf{Training Details.}}

\begin{table}[!ht]
\centering
\footnotesize
\begin{tabular}{c|ccc|ccc}
\hline
\hline
Model                    & \multicolumn{3}{c|}{Ours}     & \multicolumn{3}{c}{Baseline}  \\ \hline
\multirow{2}{*}{Testing} & \multicolumn{3}{c|}{Training} & \multicolumn{3}{c}{Training}  \\
                         & AllRecipes & FOOD.com & Both  & AllRecipes & FOOD.com & Both  \\ \hline
AllRecipes               & 96.98      & 94.38    & 97.30 & 96.31      & 92.82    & 96.16 \\
FOOD.com                 & 87.51      & 91.45    & 93.15 & 47.20      & 41.45    & 50.89 \\
Both                     & 89.62      & 92.28    & 93.64 & 57.99      & 50.98    & 60.83 \\ \hline
\hline
\end{tabular}
\caption{Experimental Results: F1 Score.}
\label{table:results1}
\end{table}
We choose the number of layers, $N$, in the recurrent module as 4. The learning rate is 5e-5 with a batch size equal to 1.
During the training, we minimize the sum of the cross-entropy losses.

\paragraph{\textbf{Results.}}

\begin{table*}[!th]
\centering
\footnotesize	
\begin{tabular}{l|ccc|ccc}
\hline
\hline
\multicolumn{1}{c|}{Model}  & \multicolumn{3}{c|}{Ours}     & \multicolumn{3}{c}{Baselines} \\ \hline
\multicolumn{1}{c|}{Entity} & Recall & Precision & F1 Score & Recall & Precision & F1 Score \\ \hline
Name                        & 92.33  & 94.31     & 93.31    & 64.54	& 64.48	& 64.51    \\
State                       & 93.41  & 94.69     & 94.04    & 48.76	& 48.50	& 48.63    \\
Unit                        & 94.65  & 96.73     & 95.68    & 62.69	& 63.69	& 63.19    \\
Quantity                    & 95.02  & 97.96     & 96.47    & 64.37	& 64.23	& 64.30    \\
Size                        & 95.10  & 93.27     & 94.17    & 32.69	& 33.33	& 33.01    \\
Temperature                 & 78.26  & 83.72     & 80.90    & 34.88	& 33.33	& 34.09    \\
Dry/Fresh                   & 93.94  & 94.90     & 94.42    & 42.35	& 43.01	& 42.67    \\
Others                      & 90.10  & 84.98     & 87.46    & 60.07	& 59.79	& 59.93    \\ 
\hline
\hline
\end{tabular}
\caption{Experimental Results at the Entity Level.}
\label{table:results2}
\end{table*}

In Table \ref{table:results1}, we present the performance of our ingredient parsing model on the AllRecipes dataset, Food.com dataset, and their combination, as shown in Table~\ref{table:dataset_statistics}. Not surprisingly, the model performance is better across the three test sets when we combine the datasets together. 
%
In Table~\ref{table:results2}, we only train and test our model on the combined dataset. 
Most entities could reach over 0.90 F1-score except for the "Temperature" and "Others" entities. Our reproduced results of the baseline model are also included and our model achieves state-of-the-art performance with a substantial improvement.
\section{Conclusion}
In this paper, we proposed a novel model to parse ingredient phrases effectively. Experimental results demonstrate that it can achieve state-of-the-art performance, over 0.93 F1-score, on the AllRecipes and Food.com datasets.


\begin{footnotesize}
\bibliographystyle{unsrt}
\bibliography{main}
\end{footnotesize}


\end{document}